# Separating Self-Expression and Visual Content in Hashtag Supervision

Andreas Veit[*]
Cornell University
andreas@cs.cornell.edu

Maximilian Nickel
Facebook AI Research
maxn@fb.com

Serge Belongie
Cornell University
sjb344@cornell.edu

Laurens van der Maaten
Facebook AI Research
lvdmaaten@fb.com

## Abstract

*The variety, abundance, and structured nature of hashtags make them an interesting data source for training vision models. For instance, hashtags have the potential to significantly reduce the problem of manual supervision and annotation when learning vision models for a large number of concepts. However, a key challenge when learning from hashtags is that they are inherently subjective because they are provided by users as a form of self-expression. As a consequence, hashtags may have synonyms (different hashtags referring to the same visual content) and may be ambiguous (the same hashtag referring to different visual content). These challenges limit the effectiveness of approaches that simply treat hashtags as image-label pairs. This paper presents an approach that extends upon modeling simple image-label pairs by modeling the joint distribution of images, hashtags, and users. We demonstrate the efficacy of such approaches in image tagging and retrieval experiments, and show how the joint model can be used to perform user-conditional retrieval and tagging.*

## 1. Introduction

Convolutional networks have shown great success on image-classification tasks involving a small number of classes (1000s). An increasingly important question is how this success can be extended to tasks that require the recognition of a larger variety of visual content. An important obstacle to increasing variety is that successful recognition of the long tail of visual content [11] may require manual annotation of hundreds of millions of images into hundreds of thousands of classes, which is difficult and time-consuming.

Images annotated with hashtags provide an interesting alternative source of training data because: (1) they are available in great abundance, and (2) they describe the long tail of visual content that we would like to recognize. Furthermore, hashtags appear in the sweet spot between capturing much of the rich information contained in natural lan-guage descriptions [15] whilst being nearly as structured as image labels in datasets like ImageNet.

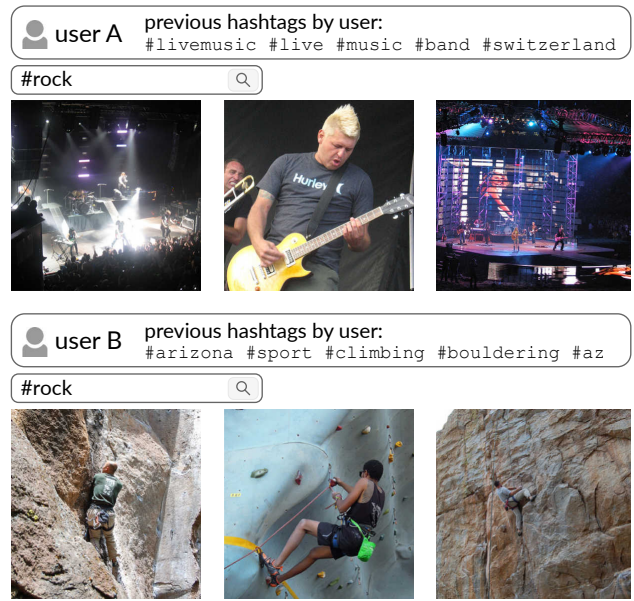

Figure 1. Image-retrieval results obtained using our user-specific hashtag model. The box above the query shows hashtags frequently used by the user in the past. Hashtag usage varies widely among users because they are a means of self-expression, not just a description of visual content. By modeling the joint distribution of users, hashtags, and images, our model disambiguates the query for a specific user. We refer the reader to the supplementary material for license information on the photos.

However, using hashtags as supervision comes with its own set of challenges. In addition to the missing-label problem that hampers many datasets with multi-label annotations (*e.g.*, [4, 13, 17]), hashtag supervision has the problem that *hashtags are inherently subjective*. Since hashtags are provided by users as a form of self-expression, some users may be using different hashtags to describe the same content (synonyms), whereas other users may be using the same hashtag to describe very different content (ambiguities). As a result, hashtags cannot be treated as oracle descriptions of the visual content of an image, but must be viewed as user-dependent descriptions of that content.

---
[*]This work was performed while Andreas Veit was at Facebook.

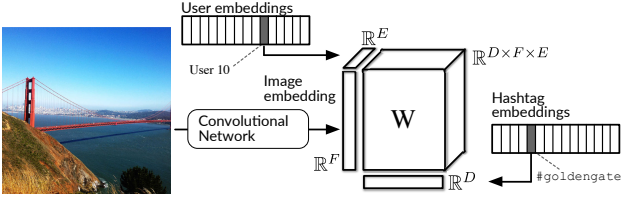

Figure 2. Overview of the proposed user-specific hashtag model. The three-way tensor product models the interactions between image features, hashtag embeddings, and user embeddings.

Motivated by this observation, we develop a *user-specific hashtag model* that takes the hashtag usage patterns of a user into account [30]. Instead of training on simple image-hashtag pairs, we train our model on image-user-hashtag triplets. This allows our model to learn patterns in the hashtag usage of a particular user, and to disambiguate the learning signal. After training, our model can perform a kind of intent determination that personalizes image retrieval and tagging. This allows us to retrieve more relevant images and hashtags for a particular user. Figure 1 demonstrates how our user-specific hashtag model can disambiguate the ambiguous #rock hashtag by modeling the user.

Figure 2 provides an overview of this model. It is comprised of a convolutional network that feeds image features into a three-way tensor model, which is responsible for modeling the interaction between image features, hashtag embeddings, and an embedding that represents the user. When multiplying the three-way interaction tensor by a user embedding, we obtain a user-specific bilinear model. This personalized bilinear mapping between images and hashtags can take into account user-specific hashtag usage patterns. Our model can produce a single score for an image-hashtag-user triplet; we use this score in a ranking loss in order to learn parameters that discriminate between observed and unobserved triplets. The user embeddings are learned jointly with the weights of the three-way tensor model.

We investigate the efficacy of our models in (user-specific) image tagging and image retrieval experiments on the YFCC100M dataset [26]. We demonstrate that: (1) we can learn to recognize large sets of visual concepts ranging from simple shapes to specific instances such as famous personalities and architectural landmarks by using hashtags as supervision; (2) our models successfully learn to discriminate synonyms and resolve hashtag ambiguities; and (3) we can improve accuracy on tasks such as image tagging by taking the user that uploaded the photo into account.

## 2. Related Work

Our study is related to prior work on (1) hashtag prediction and recommendation, (2) large-scale weakly supervised training, and (3) three-way tensor models.

Several prior works have studied **hashtag prediction and recommendation** for text posts [7, 23], infographics [2], and images [5, 31]. The most closely related to our study is [5], which studies hashtag prediction conditioned on image and user features. The main differences between our work and [5] are (1) that we train the convolutional network with hashtag supervision rather than ImageNet supervision and (2) that the user embeddings in our model are *learned* based on the images and the corresponding hashtags that users post, whereas [5] presumes user features are *pre-specified*. This allows us to model intent on the level of indiviual users, which helps in disambiguating hashtags.

Our hashtag-prediction study is an example of **large-scale weakly supervised training**, which has been the topic of several recent studies. Specifically, [24] trains convolutional networks on 300 million images with noisy labels and show that the resulting models transfer to a range of other vision tasks. Similarly, [12, 15] train networks on the YFCC100M dataset to predict words or n-grams in user posts from image content, and explore transfer of these models to other vision tasks. Our study differs from these prior works both in terms of the type of supervision used (hashtags rather than manual annotation or n-grams from user comments), and in terms of its final objective (hashtag prediction rather than transfer to other vision problems).

**Tensor models** have a long history in psychological data analysis [9, 27] and have increasingly been used in a wide range of machine-learning problems, including link prediction in relational and temporal graphs [6, 18], higher-order recommendation systems [20], and parameter estimation in latent variable models [1]. In computer vision, prominent examples of tensor models include the modeling of style and content [25], the joint analysis of image ensembles [29], sparse image coding [21] and gait recognition [8, 28].

## 3. Learning from Hashtags

Our goal is to train image-recognition models that can capture a large variety of visual concepts. In particular, we aim to learn from hashtags as supervisory signal. Formally, we assume access to a set of $N$ images $\mathcal{I} = \{\mathbf{I}_1, \ldots, \mathbf{I}_N\}$ with $\mathbf{I}_i \in [0, 1]^{H \times W \times C}$, a vocabulary of $K$ hashtags $\mathcal{H} = \{h_1, \ldots, h_K\}$, and a set of $U$ users $\mathcal{U} = \{u_1, \ldots, u_U\}$. Each image is associated with a unique user, and with one or more hashtags (we discard images without associated hashtags from the dataset). The resulting dataset comprises a set of $N$ triplets $\overline{\mathcal{T}}$, in which each triplet contains an image $\mathbf{I} \in \mathcal{I}$, a user $u \in \mathcal{U}$, and a hashtag set $\overline{\mathcal{H}} \subseteq \mathcal{H}$. Formally, $\overline{\mathcal{T}} = \{(\mathbf{I}_1, u_{m(1)}, \overline{\mathcal{H}}_1), \ldots, (\mathbf{I}_N, u_{m(N)}, \overline{\mathcal{H}}_N)\}$, in which $m(n)$ maps from the image/triplet index $n$ to the corresponding user index in $\{1, \ldots, U\}$.

Hashtag supervision differs from traditional image annotations in that it was not intended to objectively describe the image content, but merely to serve as a medium for self-
2

expression by the user. This self-expression leads to user-specific variation in hashtag supervision that is independent of the image content. We first study convolutional networks that are agnostic to the subjective nature of hashtags and simply treat them as image labels. Subsequently, we develop an user-specific model that explicitly incorporates the user as part of the hashtag-prediction model in order to capture variations in self-expression.

Throughout this work, we focus on two tasks: (1) a *tagging* task in which, given a query image $\mathbf{I}$, we aim to retrieve the most relevant hashtags for that image; and (2) a *retrieval* task in which, given a hashtag query $h \in \mathcal{H}$, we aim to retrieve the most relevant images for that hashtag.

### 3.1. User-Agnostic Hashtag Modeling

We investigate two approaches for training image-recognition models using user-agnostic hashtag supervision: (1) softmax multi-class classification and (2) hashtag-embedding regression [3]. In both cases, we learn an image model $f(\cdot; \theta) : [0, 1]^{H \times W \times C} \to \mathbb{R}^D$ which maps images into an $D$-dimensional embedding space. The image model $f(\cdot; \theta)$ is implemented by a residual network [10] with parameters $\theta$. In addition to the image model, we learn hashtag embeddings $\mathbf{h}_i \in \mathbb{R}^D$ for all hashtags $h_i \in \mathcal{H}$.

**Multi-Class Classification.** Several prior studies [12, 24] suggest that softmax classification can be very effective even in multi-label settings with large numbers of classes such as ours. Motivated by this, we train $f(\cdot; \theta)$ with a softmax over the $100{,}000$ most frequent hashtags by minimizing the multi-class logistic loss. Following [12], we select a single hashtag uniformly at random from hashtag set $\overline{\mathcal{H}}_n$ as target class for each image when training the softmax model. In particular, let $\mathbf{f}_j = f(\mathbf{I}_j; \theta) \in \mathbb{R}^D$ be the image embedding, and $h_i \in \overline{\mathcal{H}}_j$ the randomly selected hashtag. We then learn jointly the embeddings $\mathbf{h}_i$ and the parameters $\theta$ of the vision model $f(\cdot; \theta)$ by minimizing the negative log-likelihood for the probability distribution:

$$P(h_i | I_j) = \frac{\exp(\mathbf{h}_i^\top \mathbf{f}_j)}{\sum_\ell \exp(\mathbf{h}_\ell^\top \mathbf{f}_j)}. \quad (1)$$

**Hashtag-Embedding Regression.** This training method comprises two main stages. First, we learn an embedding $\mathbf{h}_i \in \mathbb{R}^D$ for each hashtag $h_i \in \mathcal{H}$. Second, we follow [3] and learn the parameters $\theta$ of the image model $f(\cdot; \theta)$ by minimizing the negative cosine similarity between the image embedding, $\mathbf{f}_j = f(\mathbf{I}_j; \theta) \in \mathbb{R}^D$, and the sum of the embeddings of the hashtags, $\overline{\mathbf{h}}_j$, corresponding to image $\mathbf{I}_j$:

$$\ell(\mathbf{f}_j, \overline{\mathbf{h}}_j; \theta) = -\frac{\overline{\mathbf{h}}_j^\top \mathbf{f}_j}{\|\overline{\mathbf{h}}_j\| \|\mathbf{f}_j\|}. \quad (2)$$

A potential advantage of this approach is that the embeddings of synonomous hashtags are likely very similar: this implies that the loss used for training the convolutional network, in contrast to the multi-class logistic loss, does not substantially penalize predicting a synonymous hashtag that the user did not happen to use to describe the image.

We experiment with two methods for learning the hashtag embeddings $\mathbf{h}_i$. The first method computes the $D$ principal singular vectors of the positive pointwise mutual information (PPMI) matrix [14]. The second method [16] explicitly models ambiguous hashtags (*i.e.*, hashtags with multiple meanings) by learning multi-sense hashtag embeddings. We follow [16] and use the global embedding vectors in their model as hashtag embedding in (2). We train all models using mini-batch stochastic gradient descent (SGD).

### 3.2. User-Specific Hashtag Modeling

The models described above do not explicitly capture variations in hashtag labels that are due to variations in how users self-express. Here, we present a model that aims to capture these variations by modeling the joint distribution of images, hashtags, and users. We will show that this can help in disambiguating the meaning of hashtags assigned to images. As before, the model represents images via a convolutional network, $\mathbf{f}_j = f(\mathbf{I}_j; \theta) \in \mathbb{R}^F$, and hashtags via embeddings $\mathbf{h}_i \in \mathbb{R}^D$. In addition, we learn user embeddings, $\mathbf{u}_k \in \mathbb{R}^E$. We aim to learn a scoring function $s(t; \mathbf{W})$ with parameters $\mathbf{W} \in \mathbb{R}^{D \times F \times E}$ that combines all three representations to predict whether or not an image-hashtag-user triplet $t$ is correct. Specifically, we select a hashtag $h_i$ from hashtag set $\overline{\mathcal{H}}_j$ uniformly at random, and model the score of the resulting triplet $t = (\mathbf{I}_j, u_k, h_i)$ as:

$$s(t; \mathbf{W}) = \sum_{r_1=1}^D \sum_{r_2=1}^F \sum_{r_3=1}^E w_{r_1 r_2 r_3} h_{i r_1} f_{j r_2} u_{k r_3}, \quad (3)$$

where $w_{r_1 r_2 r_3}$, $h_{i r_1}$, $f_{j r_2}$, and $u_{k r_3}$ are elements from $\mathbf{W}$, $\mathbf{h}_i$, $\mathbf{f}_j$, and $\mathbf{u}_k$, respectively. Equation 3 is a three-way tensor product between the embeddings of the image, hashtag, and user in which the weights $w_{r_1 r_2 r_3}$ specify the (positive or negative) interactions of all possible feature combinations. The user-specific aspect of Equation 3 can be observed by considering the summation over the user dimension. In particular, when summing over the user dimension, weighted by the embedding for user $u_k$, we obtain a user-specific weight matrix $\mathbf{W}^{(k)} \in \mathbb{R}^{D \times F}$ with entries:

$$w_{ab}^{(k)} = \sum_{r=1}^E u_{kr} w_{abr}. \quad (4)$$

The score function of Equation 3 is then equivalent to:

$$s(t; \mathbf{W}) = \mathbf{h}_i^\top \mathbf{W}^{(k)} \mathbf{f}_j. \quad (5)$$

Hence, our proposed model learns user-conditioned bilinear models between hashtags for images, by conditioning the weight matrix of the bilinear model on the user embedding.

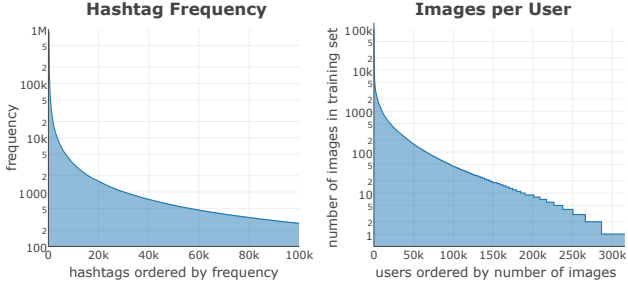

Figure 3. **Left:** Frequency of hashtags in hashtag vocabulary $\mathcal{H}$. **Right:** Number of photos per user in user set $\mathcal{U}$.

Table 1. Frequency of the most common hashtags in the data set.

| Hashtag | Frequency | Unique Users |
|---|---|---|
| `#california` | 905,715 | 15,785 |
| `#travel` | 826,366 | 15,944 |
| `#usa` | 825,641 | 13,400 |
| `#london` | 764,277 | 21,516 |
| `#japan` | 732,859 | 11,652 |
| `#france` | 650,436 | 17,265 |
| `#wedding` | 580,605 | 19,599 |
| `#music` | 552,645 | 23,359 |
| `#beach` | 547,038 | 44,695 |

Given a dataset of $M$ triplets[1] $\mathcal{T} = \{t_1, \ldots, t_M\}$, we estimate the parameters $\mathbf{W}$ using a ranking approach. In particular, we want the score of a true observed triplet $t^+ \in \mathcal{T}$ to be higher than that of an unobserved triplet $t^- \notin \mathcal{T}$. We achieve this by minimizing the following loss:

$$\ell(t^+; \mathbf{W}) = \max\left(0, \max_{t^- \notin \mathcal{T}} s(t^-; \mathbf{W}) - s(t^+; \mathbf{W}) + 1\right).$$

This ranking loss is better suited for our problem than a per-triplet binary logistic loss, because the latter would consider any unobserved triplet as a "negative". This is problematic because (1) the hashtag annotations for an image are generally not exhaustive and (2) there are far more unobserved than observed triplets. The ranking loss only aims to assign a lower score to unobserved triplets, and as a result it is not nearly as much affected by these problems.

In practice, the maximization over negative triplets $t^-$ can only be approximated. For our ranking loss to be effective, it is essential to develop good approximations for the maximization by mining "hard negatives" [22]. We adopt an online negative mining method that samples six negative triplets per positive sample, and uses each of them as a negative in the loss. Specifically, we sample three "intermediate" and three "hard" negatives. In an "intermediate" negative, one of the three elements (the image, hashtag, or user) of the positive triplet is replaced by another element that is selected uniformly at random from the training batch; the other two elements remain the same. In a "hard" negative, we replace one of the three elements in the triplet by the (non-identical) element in the training batch that maximizes the score $s(t; \mathbf{W})$.

As before, we train our user-specific hashtag model using mini-batch SGD. We first learn the parameters of the convolutional network, $\theta$, by minimizing one of the losses from 3.1. We then learn the parameters of the scoring function, $\mathbf{W}$, in a subsequent training stage. In our experiments, we use image and hashtag embeddings with 300 dimensions and user embeddings of size 50.

---
[1]Please note that $\mathcal{T}$ contains image-*hashtag*-user triplets, whereas $\overline{\mathcal{T}}$ contains image-*hashtag set*-user triplets.

Once we have inferred the embeddings for users, hashtags, and images as well as $\mathbf{W}$, we can then approach the aforementioned image tagging and retrieval results in the following way. Given a user $u_k$ and an image $\mathbf{I}_j$, we compute the most likely hashtag according to our model as:

$$\arg\max_{h_i \in \mathcal{H}} \mathbf{h}_i^\top \mathbf{W}^{(k)} \mathbf{f}_j \qquad (6)$$

The most likely image given a hashtag-user pair can be retrieved analogously.

## 4. Experiments

The aim of our experiments is: (1) to compare the strategies for training *user-agnostic* convolutional networks using hashtag supervision introduced in Section 3.1 and (2) to investigate the effectiveness of the *user-specific* hashtag model we introduced in Section 3.2.

### 4.1. Dataset

We conduct experiments on the YFCC100M dataset [26] of approximately 99.2 million photos. More than 60 million of these photos have one or more associated hashtags, and each photo has an associated user, *viz.* the user who uploaded it. We start by removing numerical hashtags and also remove the 10 most frequent tags because they are non-visual and non-informative (*e.g.*, `#iphonography`, `#instagram`, `#square`, and `#canon`). We define the hashtag set $\mathcal{H}$ as the set of the 100,000 most frequent (remaining) hashtags. The left plot in Figure 3 shows the resulting hashtag frequencies, and Table 1 lists the most frequent hashtags. The hashtag distribution is heavily skewed towards a few frequent hashtags and has a long tail of less frequent tags. For example, the most frequent hashtag, `#california`, appears over 900,000 times in the training set, *i.e.*, with 1.78% of training images. The least frequent hashtags in our hashtag set $\mathcal{H}$ only appear 260 times. Another characteristic of the hashtags is that while the most frequent tags tend to be English, less frequent tags are increasingly multilingual.

We select all photos with at least one hashtag from $\mathcal{H}$ and filter out photos by "spammers", *i.e.*, by users that use more than 15 hashtags per image on average. This results in a dataset of 55.6 million images and a user set $\mathcal{U}$ with $U = 315,745$ users. As shown in the right plot in Figure 3, the number of photos per user is also heavy-tailed.

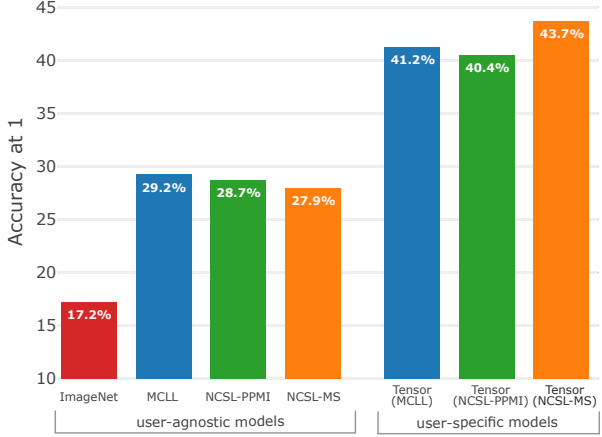

Figure 4. **Image tagging:** Accuracy@1 of four user-agnostic and three user-specific hashtag prediction models on the YFCC100M test set; see text for details. Higher is better.

Table 2. **Image tagging:** Accuracy@1 ($A$@1) and accuracy@10 ($A$@10) of two frequence baselines, four user-agnostic hashtag prediction models, and six user-specific hashtag prediction models; see text for details. Higher is better.

|   | **Method** | **A@1** | **A@10** |
|---|---|---|---|
|  | Global frequency | 1.68% | 9.65% |
|  | User-specific frequency | 38.07% | 62.55% |
| user-agnostic | Imagenet | 17.21% | 40.01% |
|  | MCLL | 29.24% | 56.47% |
|  | NCSL-PPMI | 28.72% | 47.70% |
|  | NCSL-MS | 27.94% | 46.65% |
| user-specific | MLP (MCLL) | 35.58% | 65.58% |
|  | MLP (NCSL-PPMI) | 37.31% | 67.68% |
|  | MLP (NCSL-MS) | 41.66% | 71.34% |
|  | Tensor (MCLL) | 41.24% | 70.75% |
|  | Tensor (NCSL-PPMI) | 40.43% | 68.86% |
|  | Tensor (NCSL-MS) | **43.65%** | **72.12%** |

To model a realistic use-case, we split the photos for training and testing according to their upload time stamps. We sort the photos of each user by timestamp, assign the first 90% of the images to the training set, and assign the remaining photos to the validation and test sets. This results in a training set of $N = 50.6$ million photos, a validation set of 1 million images, and a test set of 4 million images. Taken together, the dataset contains 265 million hashtags for an average of about 4.7 tags per photo.

### 4.2. Experiment 1: Hashtag Prediction

In the first set of experiments, we use our models to predict hashtags that are relevant to a given image. We measure the tagging quality of our models by their ability to predict the hashtags associated with the image in terms of accuracy@$k$ ($A$@$k$). We denote the set of the $k$ highest-scoring hashtags for image $\mathbf{I}_n$ by $\mathcal{R}_k^{(\mathbf{I}_n)}$, and as before, denote the set of hashtags that are associated to that image by $\overline{\mathcal{H}}_n$. Accuracy@$k$ is then defined as:

$$A@k = \frac{1}{N} \sum_{n=1}^{N} \frac{\mathbb{I}\left[\mathcal{R}_k^{(\mathbf{I})_n} \cap \overline{\mathcal{H}}_n \neq \emptyset\right]}{N}. \tag{7}$$

We evaluate accuracy at $k = 1$ and $k = 10$ to measure (1) how often the top-ranked hashtag is in the ground-truth hashtag set and (2) how often at least one of the the ground-truth hashtags appears in the 10 highest-ranked predictions. A key challenge in this task is that different users assign different hashtags to similar visual content: ideally, tagging methods assign hashtags that are relevant to the image content *and* are of importance to the user under consideration.

In addition to the user-specific model of Section 3.2, we evaluate four user-agnostic models: (1) a baseline model that trains a linear logistic regressor on features extracted by an convolutional network trained on ImageNet (**ImageNet**); (2) a network that is trained end-to-end for hashtag prediction using multi-class logistic loss (**MCLL**); (3) an end-to-end trained network that uses PPMI hashtag embeddings [14] in the negative cosine similarity loss of Equation 2 (**NCSL-PPMI**); and (4) an end-to-end trained network that uses the same loss but employs multi-sense hashtag embeddings [16] (**NCSL-MS**). In all experiments, our convolutional network is a ResNet-50. We evaluate three user-specific models that share the same architecture and training approach, but that vary in terms of the convolutional network that feeds image features into the three-way tensor model (those three networks were trained using MCLL, NCSL-PPMI, and NCSL-MS, respectively).

Figure 4 presents the tagging accuracy@1 of our four user-agnostic models three user-specific models on the test set. Additionally, Table 2 presents the accuracy@10 of these models, and three additional baselines: (1) a frequency baseline that predicts tags according to their frequency in the training set; (2) a *user-specific* frequency baseline that predicts tags according to their frequency for the user under consideration; and (3) a series of *user-specific* models in which we concatenate the embeddings of the three modalities and score them using a multi-layer perceptron (**MLP**) rather than the three-way tensor model.

From the results presented in the figure and the table, we make five main observations. First, all models clearly outperform the (global) frequency baseline and generally per-form quite well given that each image can be assigned one of $100,000$ different hashtags. Second, the results show that training networks from scratch for hashtag prediction substantially outperforms Imagenet-trained networks, suggesting that the visual variety in ImageNet does not suffice for hashtag prediction. Third, the user-agnostic model that was trained using multi-class logistic loss (MCLL) outperforms user-agnostic trained using negative cosine similarity loss (NCSL), in particular, in terms of accuracy at 10. Fourth, all user-specific models significantly outperform the user-agnostic models, which demonstrates the ability of these models to capture user-specific features in their predictions. Fifth, the three-way tensor models substantially outperform the user-specific frequency baseline and generally outperform the user-specific MLP baselines models, which suggests three-way tensor models are best suited for tailoring predictions based on visual content to a particular user. The highest accuracy is obtained by a three-way tensor model on top of a convolutional network trained using NCSL-MS, which is surprising because that network has the lowest accuracy of the user-agnostic models.

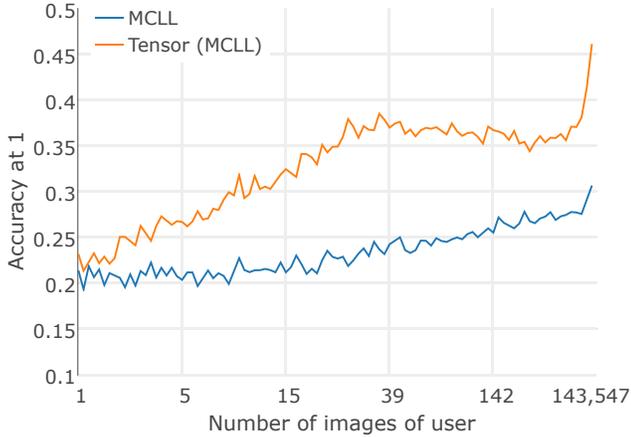

Figure 5. **Image tagging:** Accuracy@1 ($A$@1) of user-agnostic and respective user-specific tensor model as a function of the number of images by the user.

In Figure 5, we break down the tagging accuracy of our models per user by measuring accuracy as a function of the number of training images the models observed for that user. We show the accuracy break-down for the best performing user-agnostic model (MCLL) and its corresponding tensor model. The figure shows that the user-agnostic model works well across all users, but tends to perform better for users with large image libraries. We surmise this effect is due to the fact that those users have provided the majority of the images in our training set, as a result of which they dominate the data distribution. For the user-specific tagging model, we observe a stronger relationship between accuracy and the number of images per user. Whilst the user-specific model outperforms the user-agnostic one for all users, the main benefits of the user-specific modeling are for users with more than approximately 27 uploaded photos. For users with many photos, the tensor model has more data that it can use to pin down the user embeddings that capture their hashtag usage patterns.

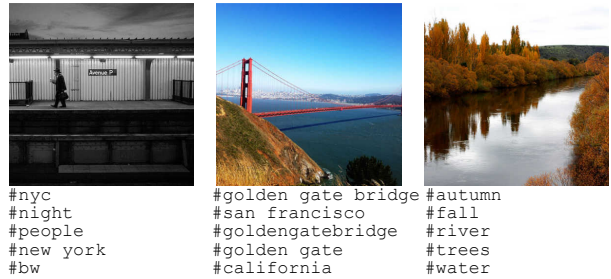

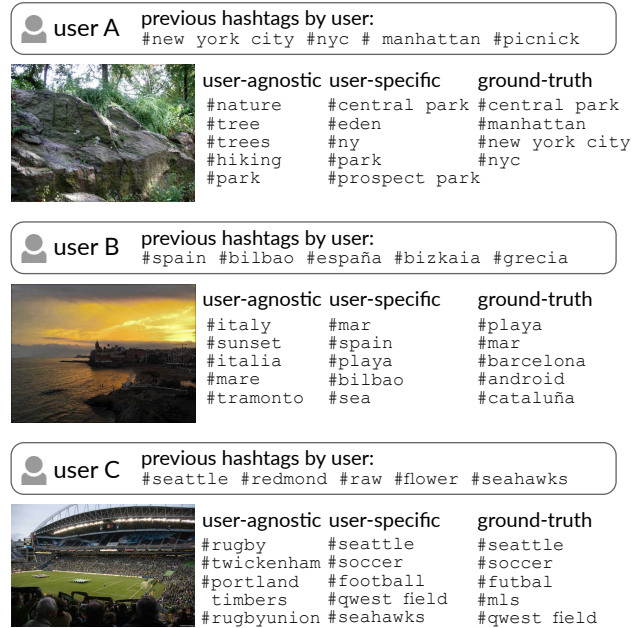

Figure 6. **Image tagging:** Example tagging results from the user-agnostic (MCLL) and user-specific (Tensor NCSL-MS) models.

Figure 6 shows examples of user-agnostic and user-specific tagging results. The tag predictions were obtained using the MCLL model and the Tensor (MCLL) model, respectively. The figure highlights the wide range of visual concepts that our convolutional networks learned to recognize. This range encompasses objects such as "people", "river", and "trees"; specific instances and locations such as the "Golden Gate Bridge", "San Francisco", and "New York"; whole-image concepts such as "autumn"; and image styles such as "black and white". The bottom part of the figure highlights the differences between the user-agnostic and user-specific models. Specifically, it shows tag predictions the user-agnostic model makes for a photo and predictions the user-specific model makes for that same photo *for a particular user* — we provide insight into the user's "profile" by showing the most frequent hashtags for that user.

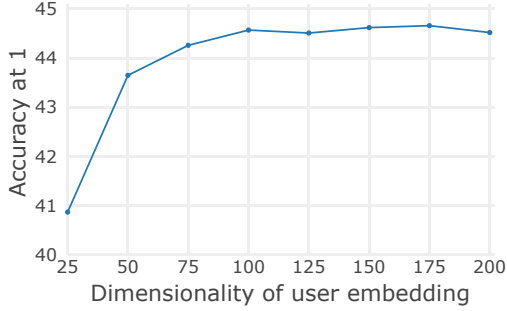

Figure 7. **Image tagging:** Accuracy@1 ($A@1$) of the Tensor (NCSL-MS) model as a function of the user embedding size, $E$.

We observe that the user-specific model can help in disambiguating (most likely) locations of a photo: *e.g.*, it changes its prediction from `#nature` to `#central park` for a user that often tags photos with concepts related to New York. The user-specific model also can change predictions into the user's preferred language (*e.g.*, from English to Spanish), and it can help in disambiguating fine-grained categories, such as recognizing the difference between a rugby and a soccer stadium. We emphasize that all the information the user-specific model used to make these disambiguations comes from image-hashtag-user triples; the model does not employ any additional user metadata.

A key to the user-specific model are the user embeddings that personalize the mapping between images and hashtags. Figure 7 shows the accuracy of the top-performing user-specific model (Tensor NCSL-MS) as a function of the dimensionality of the user embedding, $E$. The results show that a substantial number of dimensions is needed, suggesting that the user embeddings are playing an important role in the accuracy of the model.

### 4.3. Experiment 2: Hashtag-Based Image Retrieval

In a second set of experiments, we study hashtag-based image retrieval and measure the quality of our models by their ability to retrieve relevant images given a hashtag query in terms of precision@$k$ ($P@k$). We define the set of the $k$ highest-scoring images for hashtag $h$, $\mathcal{R}_k^{(h)}$, and the set of photos that are labeled with hashtag $h$, $\mathcal{GT}^{(h)} = \{\mathbf{I} \mid \exists u\colon (\mathbf{I}, u, h) \in \mathcal{T}\}$. Precision@$k$ is then defined as:

$$P@k = \frac{1}{|\mathcal{H}|} \sum_{h \in \mathcal{H}} \frac{|\mathcal{R}_k^{(h)} \cap \mathcal{GT}^{(h)}|}{k}. \tag{8}$$

We measure $P@10$ in our experiments, *i.e.*, the fraction of the 10 top-scoring images that have the query hashtag asso-

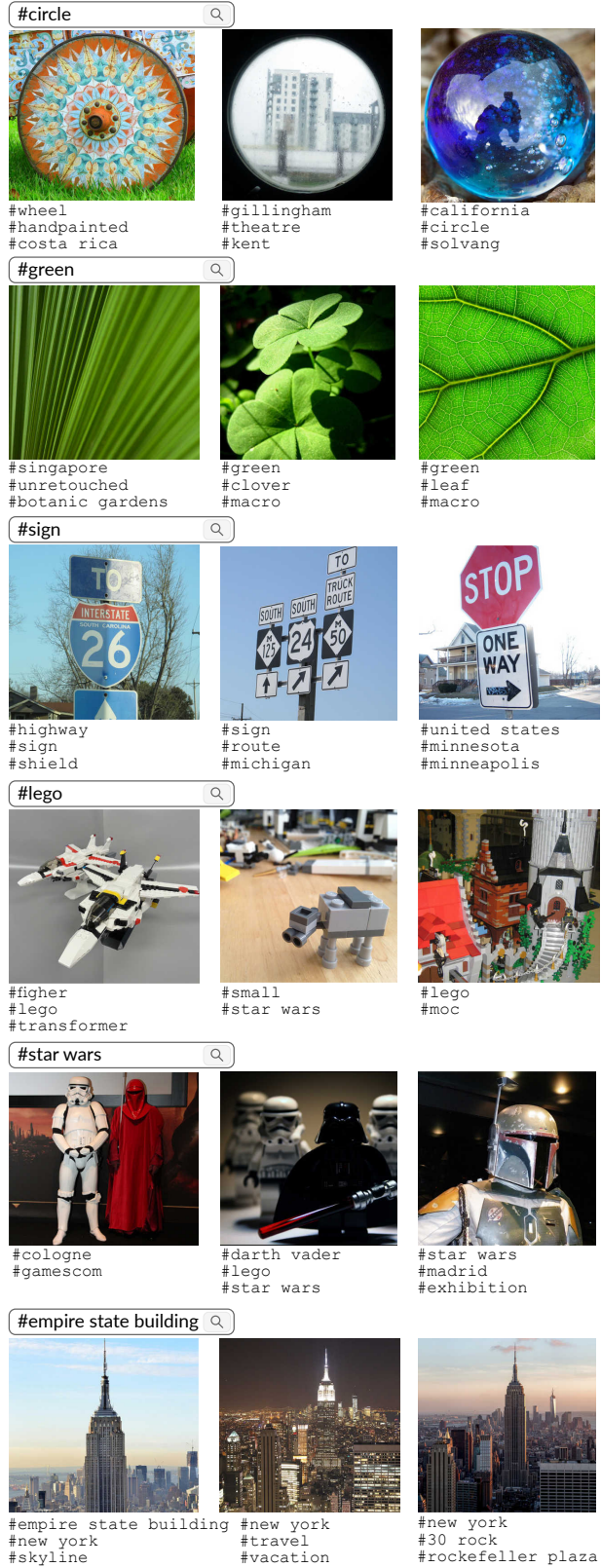

Figure 8. **Hashtag-based image retrieval:** Top-scoring photos and corresponding ground-truth hashtags for six hashtag queries. Results obtained using the user-agnostic MCLL model.

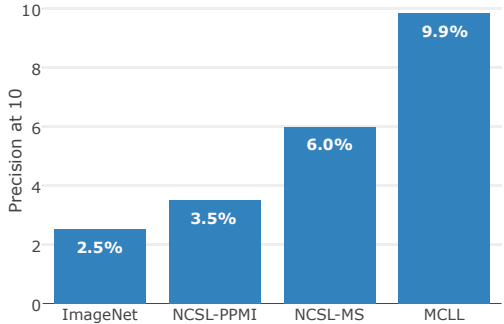

Figure 9. **Hashtag-based image retrieval:** Precision@10 ($P$@10) of four convolutional networks; see text for details. Higher is better.

ciated with it. A key challenge in this task is that hashtags can have multiple meanings: ideally, retrieval methods retrieve photos corresponding to all meanings of a hashtag.

Figure 9 presents the $P$@10 on the test set for the four user-agnostic models that were also used in Section 4.2. From the results, we make three main observations. First, similar to the first experiment, the visual variety in ImageNet does not suffice for hashtag-based image retrieval, as reflected in the low precision of the ImageNet model. Second, multi-sense embeddings (MS) seem more suitable for training with the negative cosine similarity loss (NCSL) than PPMI embeddings, presumably, because they are better at modeling ambiguous hashtags. Third, we observe that the network that was trained using multi-class logistic loss (MCLL) substantially outperforms all other models.

We emphasize that not every relevant photo for a hashtag query is also labeled with that hashtag, which gives rise to the relatively low precision values in Figure 9. We show qualitative image-retrieval results produced by the MCLL model in Figure 8, which suggest that many of the retrieved photos are actually relevant to the hashtag queries, even if they are not labeled as such. More importantly, Figure 8 illustrates the wide variety of visual concepts our models learned to recognize; the concepts recognized range from simple shapes and colors to fine-grained concepts and individual instances of architectural landmarks. Figure 1 shows an example of images retrieved by our user-specific model for the same query, #rock, for two different users. The figure demonstrates how modeling the user can help to disambiguate hashtag queries.

In Figure 10, we break down the image-retrieval precision by the frequency of the hashtags we query. The plot shows that: (1) retrieval performance is higher for frequent tags and (2) the difference between the MCLL model and the NCSL models is primarily in the long tail of less frequent tags. When evaluated on the $1,000$ most frequent tags, the classification and the multi-sense embedding model achieve a very similar precision@10 of $47\%$.

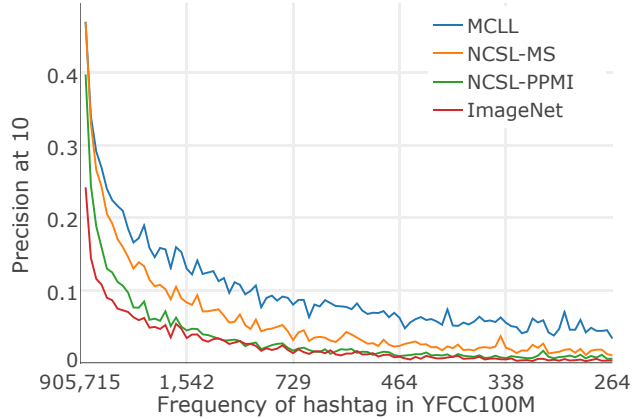

Figure 10. **Hashtag-based image retrieval:** Precision@10 ($P$@10) of four convolutional networks as a function of the frequency of the hashtag query. Higher is better.

We surmise the relatively poor performance of the embedding-regression (NCSL) models in our image-retrieval experiments is due to the hashtag embeddings being fixed in those models, whereas they are learned jointly with the visual features in the classification model. This reduces the effective capacity of embedding-regression models, resulting in weaker performances. This limitation is alleviated in the user-specific model, in which all embeddings are learned jointly. For example, we observe competitive performance of the tensor model that builds on NCSL-trained convolutional networks in the tagging experiments.

## 5. Conclusion and Future Work

This paper trained convolutional networks from scratch to perform hashtag prediction, and extended these networks with a three-way tensor model that learns user embeddings jointly with the final prediction model. This allows us to tailor the model's prediction to a specific user at test time. We used two different approaches for training the convolutional networks: a standard classification approach and an approach that regresses onto pre-learned hashtag embeddings. The classification approach performs consistently well across all tasks, whereas the embedding-regression approach mainly performs well for (user-specific) image tagging. Generally, the user-specific approach significantly outperforms the user-agnostic models demonstrating the ability to capture user-specific features in the predictions.

In future work, we intend to re-visit user-specific image retrieval in a setting in which explicit relevance information is available. Other directions for future work include incorporating user metadata [5] as well as spatial and temporal patterns [19] in our model.

# Supplementary Material for
# "Separating Self-Expression and Visual Content in Hashtag Supervision"

Andreas Veit[*]  
Cornell University  
andreas@cs.cornell.edu

Maximilian Nickel  
Facebook AI Research  
maxn@fb.com

Serge Belongie  
Cornell University  
sjb344@cornell.edu

Laurens van der Maaten  
Facebook AI Research  
lvdmaaten@fb.com

The supplementary material for the submission "Separating Self-Expression and Visual Content in Hashtag Supervision" is presented below. In Section 1, we provide license information for images from the YFCC100M dataset that are used in the main manuscript.

## 1. License Information for YFCC100M Photos

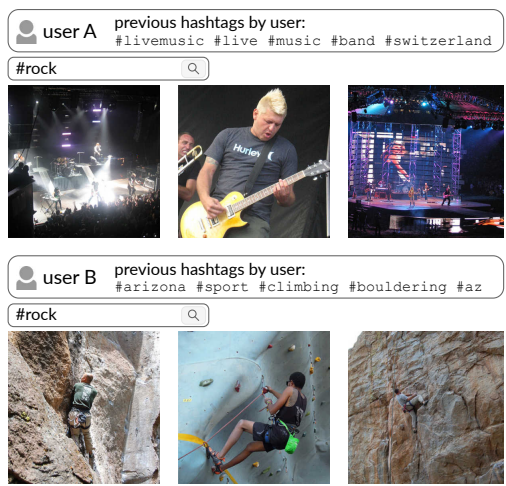

Figure 1. Image-retrieval results obtained using our user-specific hashtag model. The box above the query shows hashtags frequently used by the user in the past. Hashtag usage varies widely among users because they are a means of self-expression, not just a description of visual content. By modeling the joint distribution of users, hashtags, and images, our model disambiguates the query for a specific user. From the top-left photo in clockwise direction, the photos are courtesy of: (1) mark_donoher (CC BY 2.0); (2) joamm+tall (CC BY-SA 2.0); (3) mark_donoher (CC BY 2.0); (4) ToobyDoo (CC BY 2.0); (5) endbradley (CC BY 2.0); and (6) TonyParkin67 (CC BY 2.0).

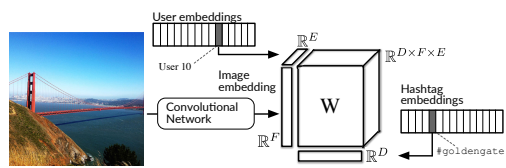

Figure 2. Overview of the proposed user-specific hashtag model. The three-way tensor product models the interactions between image features, hashtag embeddings, and user embeddings. The photo is courtesy of: Rasmus Sten (CC BY-SA 2.0).

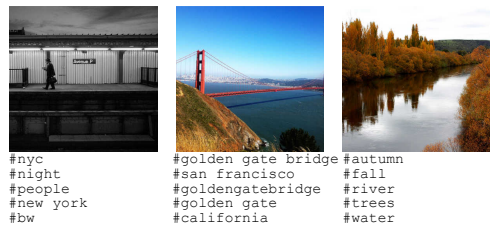
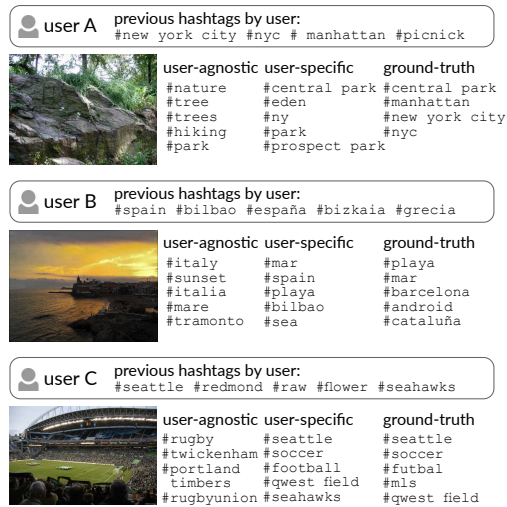

Figure 3. **Image tagging:** Example tagging results from the user-agnostic and user-specific models. For **a)** from left to right, the photos are courtesy of: (1) Leon+Fishman (CC BY 2.0); (2) Rasmus Sten (CC BY-SA 2.0); (3) Just+Emi (CC BY-SA 2.0). For **b)** from top to bottom, the photos are courtesy of: (4) Janine and Jim Eden (CC BY 2.0); (5) emubla (CC BY-SA 2.0); and (6) Luis Antonio Rodriguez Ochoa (CC BY-SA 2.0).

---

[*]This work was performed while Andreas Veit was at Facebook.

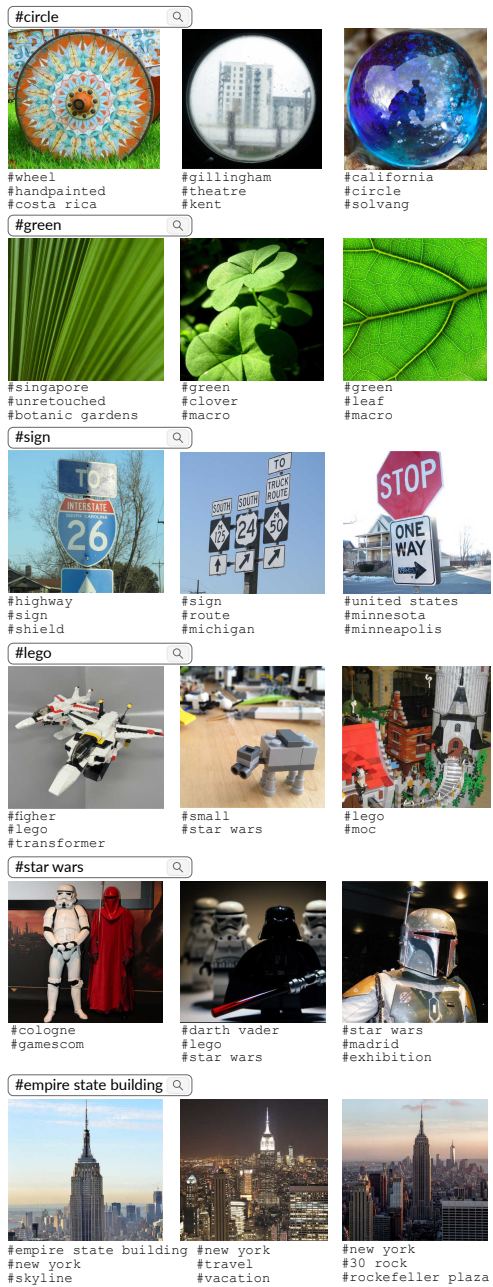

Figure 4. **Hashtag-based image retrieval:** Top-scoring photos and corresponding ground-truth hashtags for six hashtag queries. Results obtained using the user-agnostic MCLL model. Row by row from the top, from left to right within each row, the photos are courtesy of: (1) krossbow (CC BY 2.0); (2) donald+judge (CC BY 2.0); (3) ThenAndAgain (CC BY 2.0); (4) T100Timlen (CC BY 2.0); (5) mvredhed (CC BY 2.0); (6) seeks2dream (CC BY-SA 2.0); (7) Dougtone (CC BY-SA 2.0); (8) Dougtone (CC BY-SA 2.0); (9) Vandal+Tracker (CC BY-SA 2.0); (10) maxvf1 (CC BY 2.0); (11) Andrew+Turner (CC BY 2.0); (12) Timo Beil (CC BY-SA 2.0); (13) Piutus (CC BY 2.0); (14) CJ+Isherwood (CC BY-SA 2.0); (15) Tim Bartel (CC BY-SA 2.0); (16) Lola's+Big+Adventure! (CC BY 2.0); (17) Mack Male (CC BY-SA 2.0); and (18) Rick+Harris (CC BY-SA 2.0).

11